# Robust Real-Time Pedestrian Detection on Embedded Devices

Mohamed Afifi[*], Yara Ali[*], Karim Amer, Mahmoud Shaker, and Mohamed Elhelw
Center for Informatics Science, Nile University, Giza, Egypt

## ABSTRACT

Detection of pedestrians on embedded devices, such as those on-board of robots and drones, has many applications including road intersection monitoring, security, crowd monitoring and surveillance, to name a few. However, the problem can be challenging due to continuously-changing camera viewpoint and varying object appearances as well as the need for lightweight algorithms suitable for embedded systems. This paper proposes a robust framework for pedestrian detection in many footages. The framework performs fine and coarse detections on different image regions and exploits temporal and spatial characteristics to attain enhanced accuracy and real time performance on embedded boards. The framework uses the Yolo-v3 object detection [1] as its backbone detector and runs on the Nvidia Jetson TX2 embedded board, however other detectors and/or boards can be used as well. The performance of the framework is demonstrated on two established datasets and its achievement of the second place in CVPR 2019 Embedded Real-Time Inference (ERTI) Challenge[†].

## 1. INTRODUCTION

Various deep learning architectures have been proposed since Krizhevsky et. al. [2] trained a neural network model of multiple convolutional and feedforward layers on large dataset of images for object classification, numerous deep learning architectures have been proposed. One family of these architectures is designed for object detection which entails predicting bounding boxes that enclose objects of interest in a certain image. The state of the art approaches for this task can be roughly divided into two categories. The first include two-stage models such as R-CNN [3], Fast R-CNN [4], Faster R-CNN [5] and SPP-net [6]. These models propose search regions then process and classify those regions. The second category comprises single-stage models such as Yolo [7] and SSD [8]. Two-stage object detection models achieve better accuracy but with slow inference due to demanding computations. On the other hand, single-stage models are faster with lower accuracy compared to two-stage models.

In order to deploy the above models onboard of embedded devices, two important aspects must be taken into consideration. First, typical embedded devices have limited computational power. Second, a sequence of images (i.e. video) must be processed. Recent work aimed to address these constraints by creating light-weight versions of original models such as Tiny-Yolo1 and SSD300.8 Other approaches such as MobileNet [9] and ShuffleNet [10] optimize the base of a pre-trained network to have higher FPS. Lu et. al. [11] incorporated a Long Short-Term Memory (LSTM) model to make use of the spatio-temporal relation among consecutive frames in a video while Broad et. al. [12] added a convolutional recurrent layer to the SSD architecture to fuse temporal information.

This paper proposes a novel framework for robust real-time pedestrian detection in videos captured above street level such as those from pole-mounted security cameras. The framework uses the Yolo-v3 as its backbone detector but will work with other detectors with similar features. It exploits temporal information in videos while performing real-time inference by combining deep learning models pre-trained on large scale dataset of single images. Multiple input resolutions are used to perform robust pedestrian detection with a high throughput making it suitable for real-time operation on the Nvidia Jetson TX2 and similar embedded boards. Figure 1 shows an example where the proposed framework clearly achieves improved results compared to the Yolo-v3 detector.

---

[*] Indicates equal contribution
Further author information: (Send correspondence to Karim Amer)
Karim Amer: E-mail: k.amer@nu.edu.eg
[†] https://sites.google.com/site/uavision2019/

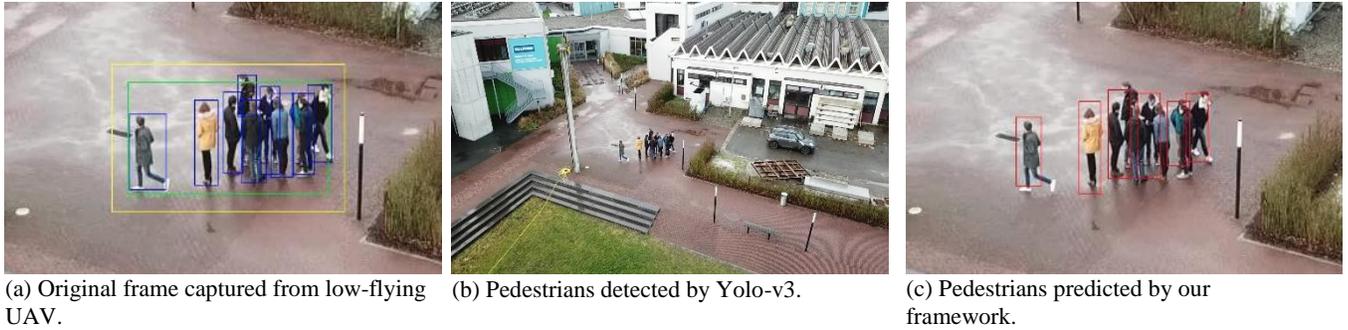

(a) Original frame captured from low-flying UAV.

(b) Pedestrians detected by Yolo-v3.

(c) Pedestrians predicted by our framework.

Figure 1. A qualitative comparison between Yolo-v3 and the proposed framework. The green box is the proposed crop. The yellow box is the crop after padding and expansion. Blue boxes are final pedestrian detections inside the crop. Images from the CVPR 2019 ERTI Challenge dataset.

The paper is organized as follows: Section 2 introduces the proposed framework, Section 3 discusses the framework workflow and details, Section 4 explains the crop proposal procedure whereas Section 5 presents results obtained in the CVPR 2019 Embedded Real-Time Inference (ERTI) Challenge and with the VisDrone dataset [13].

## 2. FRAMEWORK OVERVIEW

Robust real-time detection of pedestrians from robots and security cameras entails exploiting both temporal and spatial characteristics. Temporal relationships, however, are not utilized in many state-of-the-art lightweight detectors such as Yolo-v3 which results in objects being irregularly detected and missed in consecutive frames. Another impediment to attaining real-time performance is that most detectors need to process the full-resolution image and significant image downscaling can result in inaccurate detections. For instance, processing the full resolution image in Yolo-v3 is computationally expensive in terms of both time and memory requirements making it unsuitable for deployment on embedded devices. On the other hand, downsizing a high-resolution frame of 1920x1080 to 416x416 or 618x618 results in inaccurate bounding box positions.

Furthermore, it has been observed that in many footages captured from pole-mounted cameras, pedestrians can be clustered around certain locations in the image. In this case, reduced computations can be achieved by processing only those parts in the image with high probability to contain objects, which we call "crop regions". To this end, our framework proposes crop regions in the current frame as clusters of bounding boxes of objects that were detected in the previous frame. Consequently, when the framework processes the current frame, it focuses on those crop regions to attain more precise, i.e. fine, detections whereas for other image regions, Yolo-v3 is used but with downscaled input size and hence a decreased accuracy that improves over subsequent frames.

Temporal relationships can also be exploited to avoid missing pedestrians that were detected in the previous frame. The framework utilizes the confidence score that Yolo-v3 generates for each bounding box. Boxes with high confidence score are accepted as genuine detections whereas boxes with low confidence score are still accepted as valid detections if they have high overlap with genuine detections from previous frame.

## 3. FRAMEWORK DETAILS

The proposed framework processes a sequence of frames from a video recording or camera stream. The flow of the framework is illustrated in Figure 2. The system starts by running Yolo-v3 on the first frame to get a set of initial bounding box detections as shown in the upper part of Figure 2. These boxes are used to generate a set of large crops and another set of small crops. The set of large crops usually cover areas where large numbers of pedestrians are clustered, while small crops usually cover individual and/or small groups of pedestrians. These two sets of crops are produced by first passing the initial bounding boxes to the algorithm explained in Section 4 to generate a number of large size crops (in our case we generate 3 crops of maximum size 256x448). Subsequently, boxes that are not covered by the large crops are fed to the same crop proposing algorithm to generate a set of small crops (in our case we generate up to 20 small crops of maximum size 160x96).

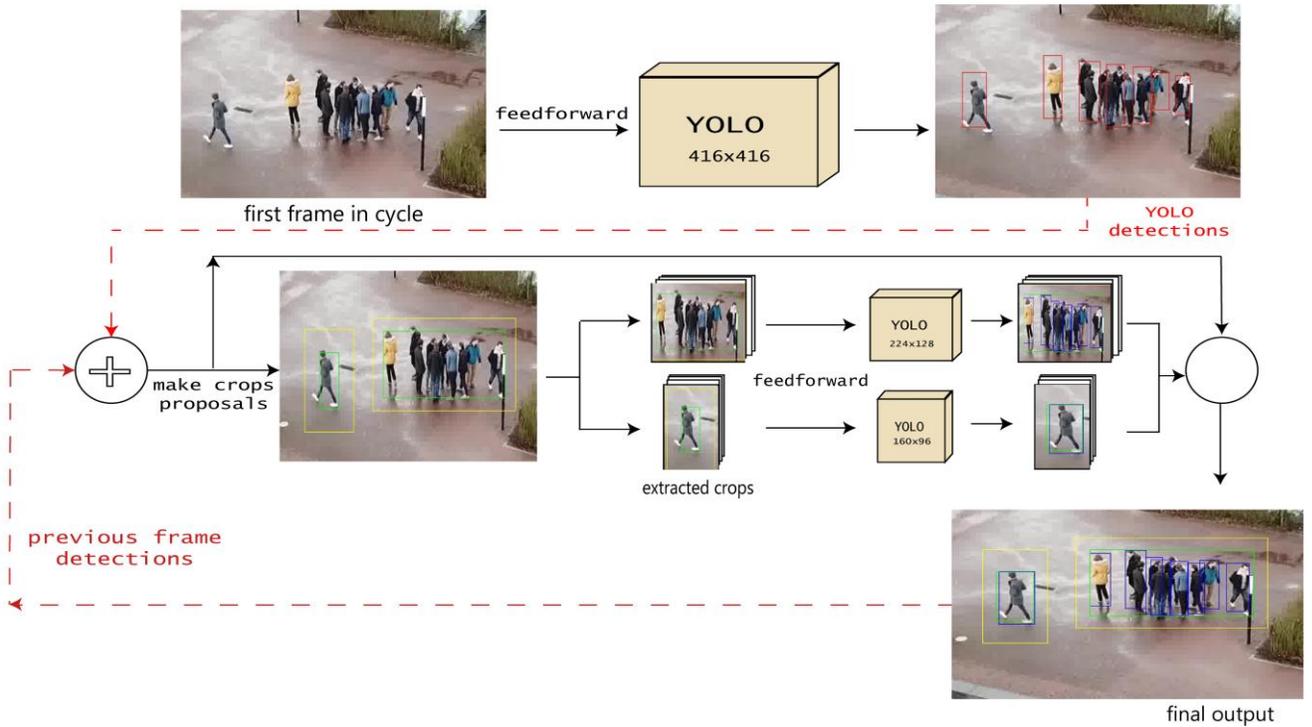

Figure 2. Proposed framework workflow

---

**Algorithm 1** Proposing Crops
---

**procedure** $ProposingCrops(B, k, W_{max}, H_{max})$
   **INITIALIZE** $E \leftarrow$ set of all edges in the undirected fully-connected graph implied by $B$

   **INITIALIZE** $forest$ such that each node is in a separate tree

   Sort $E$ in the ascending order according to edge weights

 **For each** edge $e \in E$
    **if** Number of trees in $forest \leq K$ **then return** $forest$
    $u, v \leftarrow$ The two nodes that share the edge $e$
    **if** $u$ and $v$ are already in the same tree in $forest$ **then continue**
    **if** merging trees of $u$ and $v$ gives a crop with valid dimensions **then** Merge the two trees containing
    $u$ and $v$ in $forest$ into a single tree by connecting them using the edge $e$
 **return** $forest$

---

Afterwards, each crop is resized to a predefined size according to the set that it belongs to (we resize large crops to 224x128 and small crops to 160x96). These two sets of crops are then pre-processed and passed to Yolo-v3 to get a set of predicted boxes inside each crop as illustrated in the middle part of Figure 2. Finally, temporal relation between consecutive frames is employed to make sure pedestrians detected are not dropped in intermediate frames. To this end, the confidence score that Yolo-v3 outputs for each detected box is utilized. In the current implementation, a confidence threshold of 0.2 is used to accept high confidence boxes as genuine detections and filter out low confidence boxes unless they have high overlap with genuine detections in the previous frame. This way, detections with low confidence score are still treated as valid detections. However, we do not accept any box whose confidence score is less than 0.001.

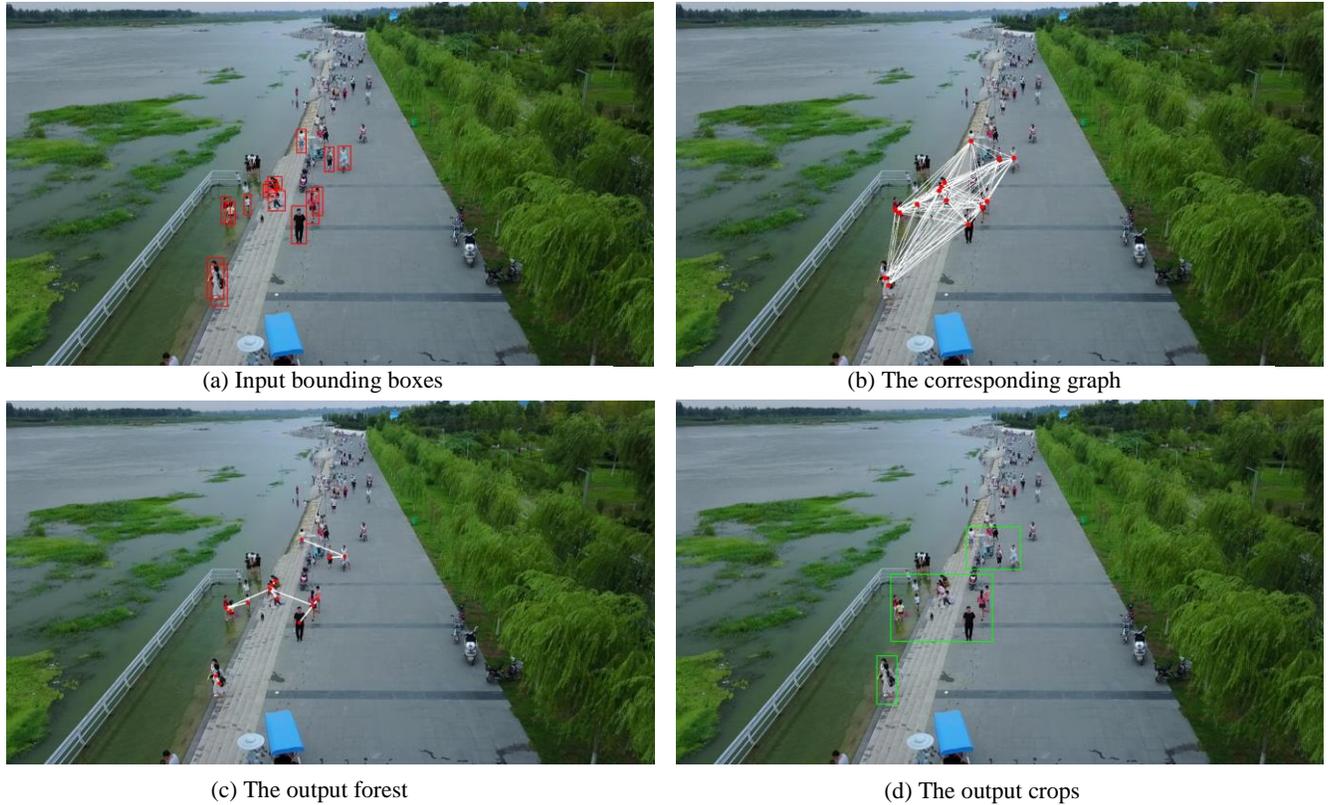

(a) Input bounding boxes  (b) The corresponding graph
(c) The output forest  (d) The output crops

Figure 3. Stages of proposing cropping regions. The image is taken from Visdrone2019 [13] training set.

The set of detected boxes in the current frame is then used to propose new crops for the next frame as indicated by the lower part of Figure 2. Yolo-v3 is applied to the entire image only once every 5 frames to detect new pedestrians who appear in the scene and update the list of crops, otherwise inference is only performed on the proposed crop regions.

## 4. CROP PROPOSAL ALGORITHM

It was observed that when the detector is provided with crops extracted from regions that contain high density of pedestrians, it produces much better detection results than when it is fed with the full frame. In the latter scenario, Yolo-v3 fails to localize most of the boxes as can be seen in Figure 1. On the other hand, extracting crops containing groups of pedestrians and feeding them to the detector produces finer boxes that better fit detected objects. The crop proposal algorithm takes as input a set of bounding boxes, a required number of crops to be generated and the maximum width and height of the crops. It outputs a set of proposed crops as shown in Figure 3d.

The crop proposal problem is formulated as a graph theory [14] problem. The set of bounding boxes is represented as a fully-connected graph with boxes as nodes and the weight of the edge connecting any two nodes as the distance between the centers of the corresponding boxes. Figure 3 shows a set of bounding boxes (a) and their corresponding graph (b). The problem of generating k crops is formulated as to find a forest that consists of at most k trees such that these k trees span all the nodes in the graph and the sum of weights of the edges in these trees is minimum. To achieve this, a simple modification on top of Kruskal's greedy algorithm for finding minimum spanning trees [14] is proposed with the details of the algorithm shown in Algorithm 1.

Table 1. ERTI challenge results.

| Model | mAP | FPS |
|---|---|---|
| Lucas Steinmann et al.[‡] | 64.48 | 5.61 |
| **Proposed Framework** | **62.14** | **6.02** |
| Dong Wang et al.[§] | 54.88 | 7.56 |
| Yolo-v3 608x352 [1] (baseline) | 49.83 | 11.67 |

Table 2. Proposed framework versus Yolo-v3 performance on four sequences of VisDrone2019 dataset.

| Model | mAP | FPS |
|---|---|---|
| YOLO v3-608 [1] | 37 | 3 |
| YOLO v3-416 [1] | 28 | 6.6 |
| **Proposed Framework** | **37** | **3.48** |

In Algorithm 1, $B$ is the set of input boxes, $k$ is the required number of crops, $W_{max}$ and $H_{max}$ are the maximum allowed width and height for the generated crops, respectively. We compute the dimensions of a crop corresponding to any tree (or subtree) by computing the dimensions of the smallest possible rectangle that can enclose all boxes spanned by the tree.

Checking whether two nodes belong to the same tree can be done efficiently using a Union-Find data structure [14]. If the loop terminates with $n > k$, only the largest $k$ trees in terms of number of nodes are selected whereas the remaining trees are discarded.

## 5. EXPERIMENTS AND RESULTS

This section describes the conducted experiments using the proposed framework on two datasets: the Embedded Real-Rime Inference challenge (ERTI) and VisDrone2019 datasets. All experiments were run on Nvidia Jetson TX2 embedded board. The clock frequencies of CPU, GPU and EMC are set to their maximum value and the power mode to Max-N (maximum performance mode). To accelerate the run-time inference of the deep learning models, Nvidia's TensorRT is used to leverage the capabilities of the onboard CUDA cores. Since the current implementation of TensorRT requires specifying the input size of the network in advance, we use 3 instances of Yolo-v3. One of input size 416x416 for inference on the entire frame and another two instance of input sizes 128x224 and 160x96 to handle the crops. For network training, Yolo-v3 pre-trained weights on COCO [15] dataset, was used with no fine tuning. Two metrics were used to measure the performance of the framework against state-of-the-art methods: mean Average Precision (mAP) for testing detection accuracy and Frames Per Second (FPS) for speed.

### 5.1 Embedded Real-Time Inference (ERTI)

ERTI challenge aims finding a model that can detect pedestrians in images acquired from a DJI Mavic drone flying at altitudes of around 15 meters. The collected data was held for private evaluation except for a 2 min sample video without any annotations which is not sufficient for training a deep neural network model. We annotated the video using the publicly available DarkLabel tool and used it as validation for our framework.

Qualitatively, we can see our framework has improved the detection results of Yolo-v3 in case of crowded spots as in Figure 1. Moreover, our framework is more consistent in detection because of the exploitation of temporal information as shown in Figure 4 where it detected the person on the left in both frames although Yolo-v3 dropped it in the second frame.

---

[‡] https://sites.google.com/site/uavision2019/FPDP_cameraready.pdf
[§] https://sites.google.com/site/uavision2019/dong_wang.pdf

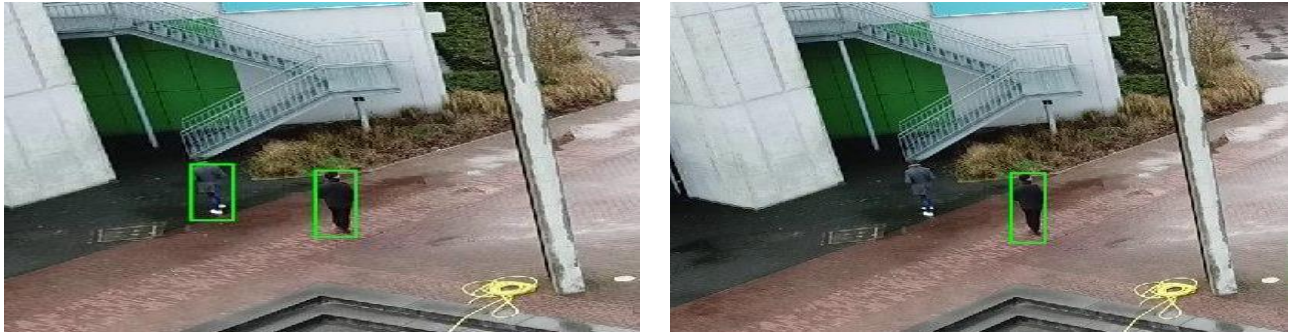
(a) The output of Yolo-v3 for two consecutive frames

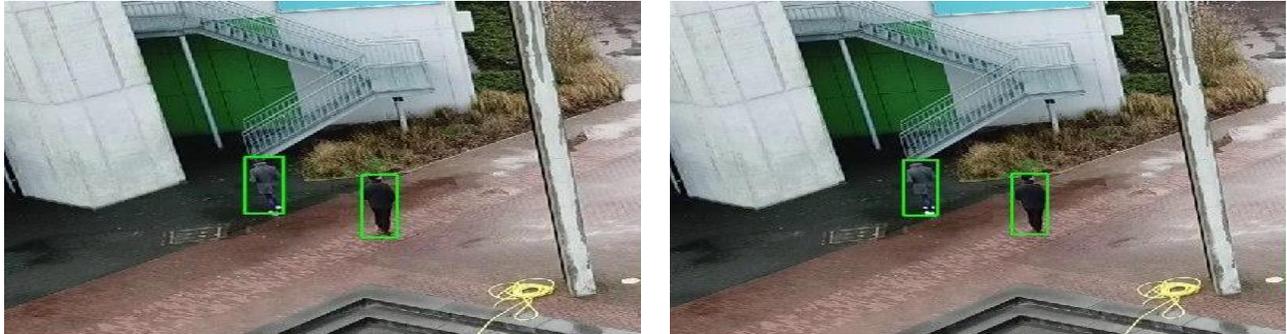
(b) The output of our framework for the same two frames

Figure 4. A qualitative example on exploiting temporal information in our framework compared to yolo-v3. The first row has yolo-v3 detection results on two consecutive frames and second row has our framework detection results on the same frames. Both frames are from CVPR 2019 ERTI Challenge dataset.

Table 1 shows the results of the best three submissions on the challenge (which are the only methods with public reports) and baseline method with Yolo-V3 608x352. Steinmann et. al.[‡] used a modified version of SSD with PeleeNet [16] backbone combined with SORT [17] tracking while Zhang et. al.[§] applied mobilenet-based SSD on different crop sizes. It can be noted that the proposed framework achieved the second place even though, unlike other competing models, our framework did not perform any finetuning.

### 5.2 Visdrone2019 Dataset

The Visdrone2019 aerial dataset [13] is used to evaluate the proposed framework. Four out of the seven Visdrone videos were selected since our framework focuses on images similar to those acquired by low flying UAVs. To this end, videos with perspective views similar to those found in the COCO [15] dataset were chosen. Furthermore, only the pedestrian class is used in the evaluation whereas other classes discarded. The four sequences that were evaluated with our framework are uav0000086_00000_v, uav0000117_02622_v, uav0000137_00458_v and uav0000339_00001_v. Table 2 compares results obtained from the proposed framework against those from YOLO416 and YOLO608. It can be seen that our framework achieves same results compared to running YOLO608-v3 over the frames of the video sequence independently while still being faster.

### 6. CONCLUSIONS

This work provides a framework for detecting pedestrians in videos captured by pole-mounted cameras and hovering robots. The framework uses the Yolo-v3 object detection network as its backbone detector applied on proposed image regions instead of the entire image. This results in higher detection accuracy while sustaining real-time performance on embedded devices such as the Nvidia Jeston TX2 embedded board. The framework alleviates the occlusion problem in crowded scenes and provides a lightweight detector that can detect pedestrians with different sizes. The temporal relation among sequential frames is exploited for robust performance.

Future work includes incorporating temporal information with compressed representations for tracking over multiple frames [18] as well as utilizing structural constraints [19] to further enhance overall detection results.